\documentclass{article}
\usepackage{iclr2027_conference,times}
\usepackage{amsmath,amssymb,graphicx,booktabs,url,xcolor}
\usepackage{nicefrac} %
\usepackage{longtable}
\usepackage{array}
\usepackage{multirow}
\usepackage{float}
\usepackage[hidelinks]{hyperref}
\usepackage[protrusion=true,expansion=true,tracking=false,kerning=false,spacing=false]{microtype}

\usepackage[noend]{algpseudocode}
\floatstyle{ruled}
\newfloat{algorithm}{t}{loa}
\floatname{algorithm}{Algorithm}
\algrenewcommand\algorithmicrequire{\textbf{Input:}}
\algrenewcommand{\alglinenumber}[1]{\scriptsize#1:}

\title{Second Thoughts: Exploring and Planning through Comparative Imagination in World Models}

\iclrfinalcopy
\author{Hieu Tran$^{1}$\quad Hong Yu$^{1,2}$\\
\textnormal{$^{1}$University of Massachusetts Amherst\quad $^{2}$University of Massachusetts Lowell}\\
\textnormal{\texttt{hieutran@umass.edu}}}

\begin{document}
\maketitle
\lhead{Preprint}

  \begin{abstract}
  World models let agents compare imagined outcomes before acting, but an agent
  must still decide which alternatives to explore and when a proposed plan should
  replace its learned policy. We introduce \textbf{SDI}(\textbf{S}ibling \textbf{D}isagreement in \textbf{I}magination), which connects these decisions through a shared actor-based reference.
  \textbf{SDI} compares action sequences imagined from the same model state, using repeated
  rollouts of the actor's own sequence to measure variation under fixed behavior.
  This reference scales both the spread across candidate returns, which provides
  an exploration signal, and a proposed plan's predicted advantage, which governs
  action acceptance. The agent executes a planned action when its advantage exceeds
  the reference variability by a fixed factor; otherwise, it follows the actor.
  \textbf{SDI} reuses the base agent's world model, actor, and critic without requiring an
  additional disagreement ensemble. Exploration and planning guide training data
  collection, while evaluation uses the learned actor alone. Across
  Meta-World, Adroit, and
  DM-Control, \textbf{SDI} achieves a higher observed
  task-average score than every compared baseline in each suite and matches or
  exceeds the strongest baseline's performance on $15$ of $18$ tasks.
  Component ablations and robustness analyses further examine the roles of
  exploration, planning exposure, and action acceptance. Paired evaluations from
  identical simulator states show that the gate improves both recovery from actor
  failures and preservation of successful behavior relative to ungated planning,
  while substantially reducing return degradation. Together, these results support
  comparing imagined alternatives against a shared actor reference as a unified approach
  to exploring and planning with world models.\footnote{\label{fn:code}Code and configurations are available in the anonymous repository: \url{https://anonymous.4open.science/r/sdi-EEDA}.}
  \end{abstract}

\begin{figure}[t]\centering
\includegraphics[width=\linewidth]{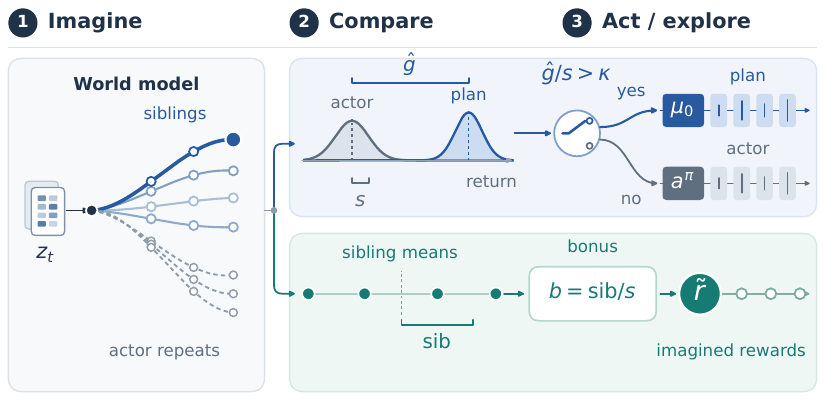}
\caption{\textbf{Overview of \textsc{SDI}.}
\textbf{Imagine:} candidate action sequences (siblings, blue) and repeated rollouts
of the actor's fixed sequence (dashed gray) start at the same latent state $z_t$.
\textbf{Compare:} fresh plan and actor rollouts provide the mean return gap $\hat g$
and actor-return standard deviation $s$; dots represent candidate mean returns,
whose spread is measured by $\mathrm{sib}$.
\textbf{Act/explore:} the upper branch selects the planned action $\mu_0$ or actor
fallback $a^\pi$; dark cells indicate the executed action. The lower branch normalizes,
clips, and weights the exploration bonus $b$ to augment the first imagined reward
$\tilde r$. The bonus anneals after the first training success. Exploration and gating
use separate rollout settings; evaluation uses the actor alone.}
\label{fig:schematic}
\end{figure}

\section{Introduction}
Learned world models support two complementary ways to improve control. Agents such as
Dreamer train a fast actor on imagined experience \citep{hafner2020dreamer,hafner2025nature},
whereas PlaNet and TD-MPC2 search through predicted futures when an action is needed
\citep{hafner2019planet,hansen2024tdmpc2}. Hybrid approaches use a learned actor or
value function to guide online search \citep{sikchi2021loop}. These approaches make
imagination useful both for learning a policy and for revising its decisions. An actor
acts quickly, whereas a planner searches at each step. During online learning, planned
actions also change the actor's training data. When learning stalls, when should a
searched action replace the actor's choice?

A planner optimizes predictions from an imperfect model. Its highest-scoring sequence
can therefore exploit errors in those predictions rather than improve the real outcome
\citep{talvitie2017selfcorrect,lambert2020mismatch,sims2024edgeofreach}.
Searching many alternatives can favor prediction errors, and even an accurately
predicted high return offers little reason to intervene if the actor would do as well.
Probabilistic ensembles and short model rollouts limit some forms of this risk
\citep{chua2018pets,janner2019mbpo}, but a decision still requires comparing a proposed
plan with the action the agent would otherwise take. During training, that choice also
changes the experience from which the actor learns. It is thus important to distinguish
the effect of planning on the eventual policy from the effect of a particular
intervention \citep{hamrick2021role}.

Exploration poses a related problem. Prediction error, novelty, and model disagreement
can direct an agent toward unfamiliar experience
\citep{pathak2017icm,burda2019rnd,pathak2019disagreement,sekar2020plan2explore}.
These signals are valuable for learning the environment, but familiarity with the
dynamics and progress on a particular task are different properties. Even a predictable
state may offer actions with very different task outcomes.
When a task reward is specified, the agent may therefore seek
actions with meaningfully different predicted returns. A large spread across imagined
actions is hard to interpret on its own, however: stochastic rollouts can also disagree
about the return of one fixed action sequence. A useful comparison needs to account for
that variation while remaining relevant to the actor's current behavior.

We introduce \textbf{SDI}(\textbf{S}ibling \textbf{D}isagreement in \textbf{I}magination), which compares imagined
action sequences with repeated rollouts of the actor's fixed sequence from the same
model state. This actor-based reference guides exploration and plan acceptance during
data collection; evaluation uses the actor alone. SDI reuses the existing world model
and search optimizer. Across Meta-World~\citep{yu2020metaworld}, Adroit~\citep{rajeswaran2018dapg}, and
DM-Control~\citep{tassa2018dmc}, \textsc{SDI} matches or exceeds the strongest compared
baseline on $15$ of $18$ tasks. In a replicated, matched-occupancy ablation, it raises
mean precision-manipulation success from $32\%$ with ungated planning to $53\%$.

Our contributions are:
\begin{itemize}\setlength{\itemsep}{2pt}\setlength{\parskip}{0pt}\setlength{\topsep}{3pt}
\item \textbf{A task-directed measure of imagined disagreement.} We define sibling
disagreement as variation in predicted task returns across action sequences imagined
from the same latent state, measured relative to repeated rollouts of the actor's
fixed sequence. This comparison separates action-dependent return variation from
stochastic variation under fixed behavior.
\item \textbf{A unified rule for exploration and selective planning.} We introduce
\textsc{SDI}, which uses the same actor-repeat reference to reward informative imagined
alternatives and to accept a planned action only when its predicted advantage exceeds
local actor variability. SDI requires no additional disagreement ensemble, guides
data collection during training, and retains actor-only evaluation.
\item \textbf{Broad and intervention-level empirical evidence.} We demonstrate that
SDI achieves the highest observed task-average score among compared methods in each
of three benchmark suites and matches or exceeds the strongest baseline mean on
$15$ of $18$ tasks. Matched-occupancy ablations and paired-state executions further
show improved actor learning, recovery, success preservation, and return retention
relative to ungated planning.
\end{itemize}

\section{Related work}
\label{sec:related}
\textbf{World-model learning and planning.} Dreamer trains an actor in imagined rollouts and executes it without further search
\citep{hafner2020dreamer,hafner2021dreamerv2,hafner2025nature}; R2-Dreamer provides the primary base agent in our experiments \citep{morihira2026r2}. Decision-time planners instead search through model-based imagination \citep{hafner2019planet,schrittwieser2020muzero,ye2021efficientzero,hansen2022tdmpc,hansen2024tdmpc2,nagabandi2019pddm,dinowm2024},
while hybrid methods combine search with learned policies or values
\citep{sikchi2021loop,anthony2017exit}. LOOP combines finite-horizon model lookahead with
an off-policy terminal value and constrains search relative to an actor
\citep{sikchi2021loop}. SDI adds an explicit post-search comparison against repeated
actor rollouts and uses that same reference construction for exploration. This targets
the acceptance of a proposal as well as the experience collected through planning
\citep{hamrick2021role}.

\textbf{Model error and uncertainty.}
Model error can compound over a rollout \citep{talvitie2017selfcorrect}, prediction objectives
need not align with control \citep{lambert2020mismatch}, and planning can exploit poorly
covered regions \citep{sims2024edgeofreach}. Short rollouts, ensembles, and uncertainty-aware
updates address these problems in different ways
\citep{janner2019mbpo,chua2018pets,buckman2018steve,yu2020mopo,pan2020m2ac,frauenknecht2024macura}.
PETS propagates uncertainty through probabilistic ensembles and trajectory sampling
\citep{chua2018pets}. SDI obtains its reference from stochastic repeats within the
base world model. This estimates rollout variability for a fixed actor sequence;
model bias remains possible \citep{berger2026biased}. Our paired-state experiments
test whether a margin scaled by that reference identifies executed improvements.

\textbf{Exploration and action comparisons.}
Intrinsic rewards use prediction error, novelty, information gain, or ensemble disagreement
to encourage exploration
\citep{pathak2017icm,pathak2019disagreement,burda2019rnd,vime,sekar2020plan2explore,kauvar2023curiousreplay}.
Other signals measure action influence or reward disagreement
\citep{seitzer2021cai,bierling2025dreamerv3xp}. Adaptive mechanisms switch exploration
modes or balance intrinsic and extrinsic rewards \citep{pislar2022when,chen2022eipo},
while dormant-representation diagnostics provide a different basis for adapting training
interventions \citep{sokar2023dormant,xu2024drm}.
Plan2Explore rewards disagreement among predictions of the next observation embedding
to seek expected information gain \citep{sekar2020plan2explore}. SDI instead compares
task-return predictions across action sequences, relative to actor-repeat variability.
This makes the signal task-directed and ties it to the reference used for plan acceptance.
The construction relates to interventional advantage, counterfactual credit assignment,
and twin models \citep{pan2022dae,pan2024skillorluck,mesnard2021cca,balke1994twin,vlontzos2023twin};
interventional causal representation learning likewise studies information obtained
under interventions \citep{ahuja2023interventional}. We therefore evaluate SDI's predictions using executed alternatives.

\section{Method}
\label{sec:method}
\textsc{SDI} augments a world-model actor--critic during training while leaving the
base learning objectives unchanged. At each latent state, it constructs a local actor
reference from repeated imagined rollouts of a fixed actor sequence. The same reference
scales a task-directed exploration bonus and the advantage required to execute a planned
action. Planning changes the experience collected for learning; evaluation executes the
learned actor's mode without planning. Algorithm~\ref{alg:sdi} summarizes the complete
training-time interaction.

\subsection{Actor-referenced imagined returns}
\label{sec:iu}
The base agent provides a recurrent latent dynamics model
$z_{\tau+1}=f_{\mathrm{img}}(z_\tau,a_\tau)$, reward head $r$, continuation head
$\mathrm{cont}$, actor $\pi$, and critic $V$. For an $H$-step action sequence, its
continuation-weighted imagined return is
\[
G(a_{0:H-1})=\sum_{\tau<H}d_\tau r(z_{\tau+1})+d_HV(z_H),
\qquad d_0=1,\quad d_{\tau+1}=d_\tau\gamma\,\mathrm{cont}(z_{\tau+1}).
\]
Model predictive path integral (MPPI) search in latent space \citep{mppi} evaluates candidate
action sequences by averaging stochastic model rollouts. After search, \textsc{SDI}
re-evaluates the selected plan and the actor's fixed mode sequence with $R_g$ fresh
rollouts each. Let $\overline G_{\mathrm{plan}}$ and $\overline G_\pi$ denote their
mean returns, and let $s_\pi$ denote the standard deviation across actor-sequence
rollouts. The imagined advantage and its normalized score are
\begin{equation}
\hat g(z)=\overline G_{\mathrm{plan}}-\overline G_\pi,
\qquad
\mathrm{IU}(z)=\frac{\hat g(z)}{s_\pi(z)}.
\label{eq:iu}
\end{equation}
Both sequences start from the same latent state, so their mean-return gap compares
the proposed change in behavior against what the actor would do there. The denominator
answers a different question: how much can the predicted return vary when the action
sequence is held fixed? A plan whose advantage is small relative to that local variation
provides less reason to replace the actor than the same raw gap in a stable state.
This scale does not correct model errors shared by the two sequences or turn
$\mathrm{IU}$ into a calibrated probability of improvement.
The actor sequence is generated once and held fixed across its repeated rollouts, so
$s_\pi$ measures return variation under fixed behavior. Exploration and action gating
share the actor-referenced comparison, not the sampled trajectories: exploration uses
short sibling rollouts and its own actor repeats, whereas the gate freshly re-scores a
selected plan and actor sequence over the planning horizon. All imagined rollouts use
independent model noise; Appendix~\ref{app:coupling}
distinguishes this sampling from the paired simulator randomness used only for analysis.

\begin{algorithm}[t]
\caption{\textbf{\textsc{SDI} training-time interaction.} Exploration modifies imagined
rewards, while selective planning modifies environment actions.}
\label{alg:sdi}
\begin{algorithmic}[1]
\footnotesize
\Require latent $z_t$, actor $\pi$, critic $V$, model $f_{\mathrm{img}}$
\State Draw fallback action $a_t\sim\pi(\cdot\mid z_t)$
\State Update bonus weight from the first-training-success schedule
\If{the exploration bonus is active}
\State Compute $b$ from sibling and actor-reference rollouts (Eq.~\ref{eq:disagree})
\State Weight $b$ and augment the first imagined actor--critic reward
\EndIf
\State Update planner availability from episode-return history
\If{the planner is available}
\State $\mu\leftarrow\mathrm{MPPI}(z_t;N,H,I,R)$
\State Fix the actor's mode sequence and re-score it and $\mu$ with $R_g$ fresh rollouts
\State Compute $\hat g$ and $s_\pi$ (Eq.~\ref{eq:iu})
\State \textbf{if} $\hat g > \kappa s_\pi$ \textbf{then} $a_t\leftarrow\mu_0$
\EndIf
\State Execute $a_t$, append the transition to replay, and update the base agent
\end{algorithmic}
\end{algorithm}

\begin{table}[!t]\footnotesize
\caption{\textbf{Cross-domain benchmark performance under actor-only evaluation.}
Meta-World~\citep{yu2020metaworld} and Adroit~\citep{rajeswaran2018dapg} report
success; DM-Control~\citep{tassa2018dmc} reports episode return. Compared baselines
are Plan2Explore~\citep{sekar2020plan2explore}, Curious Replay~\citep{kauvar2023curiousreplay},
TD-MPC2~\citep{hansen2024tdmpc2}, DreamerV3~\citep{hafner2025nature}, and
R2-Dreamer~\citep{morihira2026r2}. Columns show half and full training budgets, and
suite means weight tasks equally.}
\label{tab:main}
\centering\footnotesize\setlength{\tabcolsep}{2pt}\renewcommand{\arraystretch}{0.90}
\setlength{\arrayrulewidth}{0.25pt}
\begin{tabular*}{\linewidth}{@{\extracolsep{\fill}}l|cc|cc|cc|cc|cc|cc@{}}
\toprule
Task & \multicolumn{2}{c|}{Plan2Explore} & \multicolumn{2}{c|}{Curious Replay} & \multicolumn{2}{c|}{TD-MPC2} & \multicolumn{2}{c|}{DreamerV3} & \multicolumn{2}{c|}{R2-Dreamer} & \multicolumn{2}{c}{\textbf{\textsc{SDI}}} \\
\cmidrule(lr){2-3}\cmidrule(lr){4-5}\cmidrule(lr){6-7}\cmidrule(lr){8-9}\cmidrule(lr){10-11}\cmidrule(lr){12-13}
 & half & full & half  & full & half & full & half & full & half & full & half & full \\
\midrule
\multicolumn{13}{@{}l}{\makebox[0pt][l]{\emph{\textbf{Meta-World}}}}\\
box-close & .00 & .00 & .67 & .98 & .81 & .98 & .00 & .22 & .58 & .97 & \textbf{.90} & \textbf{.99} \\
sweep-into & .00 & .12 & .77 & .94 & .76 & .95 & .00 & .00 & \textbf{.90} & .97 & .79 & \textbf{.98} \\
hammer & .28 & .23 & .00 & .00 & 1.00 & .99 & .00 & .00 & 1.00 & 1.00 & .53 & 1.00 \\
bin-picking & .00 & .00 & .00 & .48 & .22 & .86 & .08 & .57 & \textbf{.48} & .61 & .01 & \textbf{.92} \\
lever-pull & .08 & .12 & .00 & .45 & .24 & .37 & .24 & .57 & .40 & .90 & \textbf{.98} & \textbf{.98} \\
peg-insert-side & .00 & .00 & .04 & .96 & .05 & .96 & .05 & .59 & \textbf{.76} & .97 & .74 & \textbf{1.00} \\
hand-insert & .00 & .01 & .02 & .34 & .01 & .04 & .00 & .00 & .04 & .13 & \textbf{.11} & \textbf{.49} \\
assembly & .00 & .00 & .01 & .00 & .00 & .01 & .02 & .09 & .00 & .00 & \textbf{.07} & \textbf{.75} \\
disassemble & .00 & .00 & .00 & .00 & .01 & .00 & .00 & .32 & .01 & .12 & .00 & \textbf{.35} \\
pick-out-of-hole & .00 & .00 & .00 & .41 & .00 & .00 & .00 & .08 & .03 & .31 & \textbf{.15} & \textbf{.96} \\
\cmidrule[0.25pt](lr){1-13}
\textbf{Mean} & .04 & .05 & .15 & .46 & .31 & .52 & .04 & .24 & .42 & .60 & \textbf{.43} & \textbf{.84} \\
\midrule
\multicolumn{13}{@{}l}{\makebox[0pt][l]{\emph{\textbf{Adroit}}}}\\
hammer-v1 & .03 & .11 & .00 & .32 & .04 & .26 & .76 & .60 & \textbf{.93} & .99 & .84 & .99 \\
door-v1 & .58 & .85 & .45 & .49 & .00 & .00 & .00 & .00 & .00 & .00 & \textbf{.72} & .85 \\
pen-v1 & .04 & .00 & .16 & .19 & .10 & .23 & .07 & .19 & .06 & .20 & \textbf{.24} & \textbf{.44} \\
\cmidrule[0.25pt](lr){1-13}
\textbf{Mean} & .22 & .32 & .20 & .33 & .05 & .16 & .28 & .26 & .33 & .40 & \textbf{.60} & \textbf{.76} \\
\midrule
\multicolumn{13}{@{}l}{\makebox[0pt][l]{\emph{\textbf{DM-Control}}}}\\
cartpole-swingup & 29 & 45 & 9.8 & 80 & 0.0 & 0.0 & 51 & 190 & 26 & 242 & \textbf{62} & \textbf{279} \\
acrobot-swingup & \textbf{8.3} & 20 & 4.2 & 23 & 2.1 & 3.9 & 8.0 & \textbf{69} & 2.0 & 5.8 & 3.7 & 16 \\
ball-in-cup-catch & 537 & \textbf{973} & 780 & 445 & 469 & 634 & \textbf{937} & 959 & 750 & 949 & 820 & 961 \\
reacher-hard & 51 & 375 & 119 & 370 & 132 & 273 & 21 & 187 & 268 & 819 & \textbf{732} & \textbf{896} \\
finger-turn-hard & 198 & 304 & 84 & 186 & \textbf{199} & 373 & 163 & \textbf{616} & 163 & 592 & 179 & 579 \\
\cmidrule[0.25pt](lr){1-13}
\textbf{Mean} & 164.7 & 343.4 & 199.4 & 220.8 & 160.4 & 256.8 & 236.0 & 404.2 & 241.8 & 521.6 & \textbf{359.3} & \textbf{546.2} \\
\bottomrule
\end{tabular*}
\end{table}

\textbf{Sibling disagreement for exploration.}
Before the first successful environment transition, \textsc{SDI} evaluates $K$ action
sequences imagined from the same latent state. We call these sequences \emph{siblings}.
With $\hat G_k$ the estimated mean return of sibling $k$, their normalized return spread is
\begin{equation}
\mathrm{sib}(z)=\mathrm{std}_k(\hat G_k),
\qquad b(z)=\frac{\mathrm{sib}(z)}{s_\pi(z)}.
\label{eq:disagree}
\end{equation}
The sibling means summarize differences among action choices, whereas $s_\pi$
summarizes variation from repeating one choice. Their ratio favors states where
imagined actions differ in task return relative to that fixed-behavior variation.
The bonus affects the actor--critic update in imagination; it is not added to the
environment reward or used as the gate's plan advantage.
We divide this score by its running mean, clip it to $[0,3]$, and add it to the first
reward of the imagined actor--critic rollout with weight $\beta=0.5$. Exploration uses
$K=5$ siblings, $K_r=3$ actor-reference repeats, and horizon $3$. The first training
transition carrying the environment's success flag starts an anneal over $20$k optimizer
updates. The environment reward itself is unchanged. The anneal gradually removes the
auxiliary signal after discovery, leaving the original task reward to guide later learning.

\subsection{Search, fresh rescoring, and plan acceptance}
\label{sec:replaces}\label{sec:when}
SDI separates proposing a plan from deciding whether to execute it. MPPI first searches
for a high-return sequence by iteratively sampling candidates and refitting its action
distribution to the elites. Because the selected sequence is optimized on these search
rollouts, their returns are not reused as evidence for acceptance. Instead, SDI
re-evaluates the selected plan and the actor's fixed mode sequence from the same latent
state using independent model noise and $R_g=4$ fresh rollouts per sequence. This
post-search comparison supplies $\hat g_t$ and $s_\pi(z_t)$ in Eq.~\ref{eq:iu}; the
complete MPPI update and initialization appear in Appendix~\ref{app:planner}.
Using the actor's repeats to set the threshold also keeps acceptance tied to the
current policy, rather than to the spread of candidates that search has already
optimized.

Scoring holds the actor's mode sequence fixed, while a rejected plan executes a fresh
actor sample. This separates the gate's repeatable reference from stochastic training
behavior.

At a planning step, the agent executes the first planned action $\mu_0$ only when the
freshly estimated advantage exceeds the actor-reference variability by a factor
$\kappa$:
\begin{equation}
a_t=\begin{cases}
\mu_0,&\hat g_t>\kappa s_\pi(z_t),\\
a_t^{\mathrm{actor}}\sim\pi(\cdot\mid z_t),&\text{otherwise},
\end{cases}
\qquad \kappa=2.
\label{eq:gate}
\end{equation}
When a plan is rejected, the agent samples a fresh actor action. Because $s_\pi$
is a rollout standard deviation rather than a standard error, $\kappa$ is an empirical
operating threshold rather than a statistical confidence level.
Only the first planned action is executed; the agent re-plans from its next latent state.
These two uses of the actor reference address different decisions. Sibling disagreement
identifies states where alternative actions have meaningfully different imagined returns;
the gate tests whether a selected plan clears the actor's local return variability.
The scores use separate rollout batches and horizons. A state can therefore be useful
to explore even when no proposed plan warrants replacing the actor at that step.
Exploration changes the actor's imagined learning signal; an accepted plan changes
the next environment transition and the experience stored in replay.

\begin{figure}[!t]\centering
\includegraphics[width=.95\linewidth]{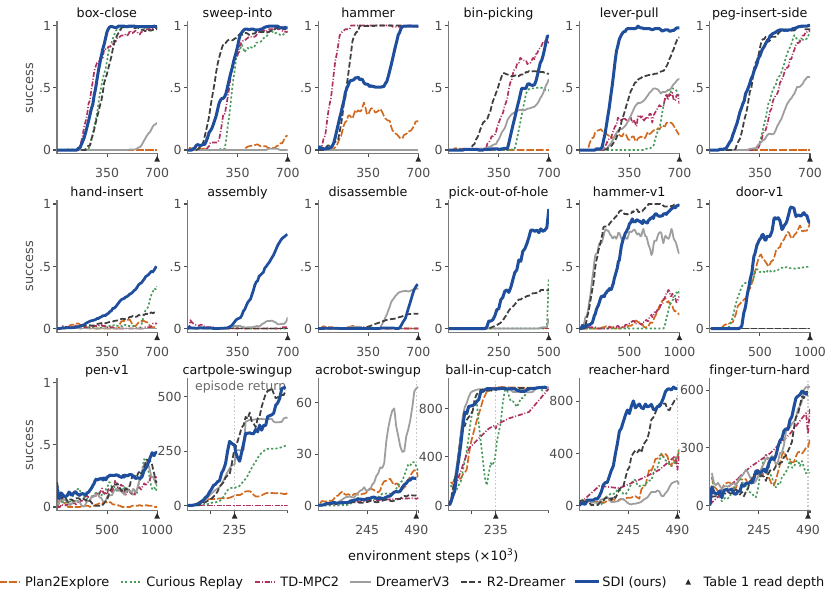}
\caption{\textbf{Cross-domain benchmark learning curves.}
Lines show average rolling success on Meta-World~\citep{yu2020metaworld} and
Adroit~\citep{rajeswaran2018dapg} and average episode return on
DM-Control~\citep{tassa2018dmc}. Black triangles mark the Table~\ref{tab:main} scoring
depths; curves for the latter suite may extend toward the nominal $500$k-step budget.}
\label{fig:curvegrid}
\end{figure}

\subsection{Training schedule and planner configuration}
\label{sec:stall}\label{sec:dose}\label{sec:uses}\label{sec:plan}\label{sec:idis}
Planning is disabled for the first $t_{\min}=20$k environment steps to allow initial
model learning. Thereafter, a return-based predicate compares an exponential average
of episode return with its value $W=50$k steps earlier and activates planning when
relative improvement falls below $.05$. Once activated, planning remains available.
We call the fraction of acting steps with planner availability \emph{occupancy}; this is
distinct from the fraction on which a planned action passes the acceptance rule.

The planner uses $N=256$ candidates, horizon $H=8$, $I=4$ refinement iterations,
$E=32$ elites, and $R=4$ stochastic rollouts per candidate. The selected plan and actor
reference are then evaluated with $R_g=4$ fresh rollouts each. These settings and
$\kappa$ are fixed across domains. Search and re-scoring occur before acceptance, so a
rejected proposal still requires the full planning computation. Appendix~\ref{app:config}
provides optimizer updates, initialization, action bounds, and variant configurations.

\section{Experimental setup}
\label{sec:setup}
\textbf{Benchmarks and budgets.}
Table~\ref{tab:main} covers three pixel-based suites.
Meta-World v3 \citep{yu2020metaworld} uses $64\times64$ observations and sparse
success rewards. The pre-specified discovery set comprises box-close, sweep-into, and
hammer, where vanilla DreamerV3~\citep{hafner2025nature} has no success by $150$k
steps. The precision-plateau set comprises hand-insert, assembly, and disassemble.
The Meta-World budget is $700$k, except for pick-out-of-hole, which uses
$500$k. Adroit \citep{rajeswaran2018dapg} includes hammer, door, and pen with dense
training rewards and success-based evaluation at $1$M steps.
DM-Control \citep{tassa2018dmc} includes five tasks at a nominal $500$k budget;
comparisons use each task's common observed depth.

\textbf{Baselines and evaluation.}
We compare with DreamerV3~\citep{hafner2025nature},
R2-Dreamer~\citep{morihira2026r2}, Plan2Explore~\citep{sekar2020plan2explore},
Curious Replay~\citep{kauvar2023curiousreplay}, and
TD-MPC2~\citep{hansen2024tdmpc2}.
All reported baseline results come from our runs. For Meta-World~\citep{yu2020metaworld}
and Adroit~\citep{rajeswaran2018dapg}, the endpoint is mean success over the final
$100$ evaluation episodes (the last $10$ evaluation calls); an episode succeeds if
its success flag is set at any step. DM-Control~\citep{tassa2018dmc} averages episode
return over five evaluation calls at a common scoring depth for each task.
\emph{Crossing} means reaching a last-$10$ evaluation success rate of at least $0.5$
at the task budget.

\begin{table}[!tbp]\footnotesize
\caption{\textbf{Ablation study on precision-manipulation tasks.}
Entries report mean evaluation success at $350$k and $700$k environment steps;
the mean column gives the equally weighted average of the three $700$k-step task means.
Occupancy is the fraction of acting steps with planner availability, not planned-action
execution. Base-agent controls are R2-Dreamer~\citep{morihira2026r2} and
DreamerV3~\citep{hafner2025nature}.}
\label{tab:balance}
\centering\footnotesize\setlength{\tabcolsep}{2pt}\renewcommand{\arraystretch}{1.10}
\begin{tabular*}{\linewidth}{@{\extracolsep{\fill}}l|l|cc|cc|cc|c|c@{}}
\toprule
Ablation & Component & \multicolumn{2}{c|}{hand-insert} & \multicolumn{2}{c|}{assembly} & \multicolumn{2}{c|}{disassemble} & \emph{mean} & \emph{occupancy} \\
\cmidrule(lr){3-4}\cmidrule(lr){5-6}\cmidrule(lr){7-8}
& & 350k & 700k & 350k & 700k & 350k & 700k & 700k & mean \\
\midrule
 & \textsc{SDI} & .11 & .49 & \textbf{.07} & \textbf{.75} & .00 & \textbf{.35} & \textbf{.53} & .97 \\
\midrule
\multirow{2}{*}{Base agent} & R2-Dreamer & .04 & .13 & .00 & .00 & .01 & .12 & .08 & .00 \\
& DreamerV3  & .05 & .16 & .00 & .00 & .00 & .00 & .05 & .37 \\
\midrule
\multirow{2}{*}{Planning} & Ungated planner & .20 & \textbf{.53} & .00 & .29 & \textbf{.03} & .14 & .32 & .97 \\
& Half occupancy & .14 & .47 & .00 & .00 & .00 & .01 & .16 & .47 \\
\midrule
\multirow{2}{*}{Exploration} & Always-on bonus & .02 & .17 & .00 & .00 & .01 & .00 & .06 & .00 \\
& Reset when stuck & .03 & .16 & .00 & .08 & .00 & .00 & .08 & .00 \\
\midrule
Replay & Spread-prioritized replay & \textbf{.27} & .40 & .00 & .00 & .00 & .00 & .13 & .47 \\
\bottomrule
\end{tabular*}
\end{table}

\section{Results}
\label{sec:results}

\subsection{Cross-domain benchmark performance}
\label{sec:main}\label{sec:cross}
\textbf{Meta-World.}
Table~\ref{tab:main} shows that across the ten Meta-World
tasks~\citep{yu2020metaworld}, \textsc{SDI} achieves the
highest observed suite mean at both the half and full training budgets, improving
from $.43$ to $.84$; the strongest baseline means are $.42$ and $.60$, respectively,
from R2-Dreamer~\citep{morihira2026r2}. At the full budget, \textsc{SDI} is highest or
tied highest on every task. Its largest gains over the strongest alternative occur
on assembly ($.75$ versus $.09$), pick-out-of-hole ($.96$ versus $.41$), and
hand-insert ($.49$ versus $.34$).

\textbf{Adroit.}
On Adroit~\citep{rajeswaran2018dapg}, \textsc{SDI} also has the highest observed suite
mean at both budgets, reaching $.60$ halfway through training and $.76$ at the full
budget, compared with $.33$ and $.40$ for the strongest baseline, R2-Dreamer. At the
full budget, it ties the best reported result on hammer ($.99$) and door ($.85$), and
leads on pen with $.44$, versus $.23$ for the strongest alternative. Thus,
\textsc{SDI} matches or exceeds the best baseline endpoint on all three tasks.

\textbf{DM-Control.}
DM-Control~\citep{tassa2018dmc} shows a more mixed task-level ranking despite the
strongest suite averages. \textsc{SDI} obtains means of $359.3$ and $546.2$ at the
half and full budgets, compared with $241.8$ and $521.6$ for the strongest baseline,
R2-Dreamer. At the full budget, it leads on cartpole-swingup ($279$) and reacher-hard
($896$), and remains close to the best result on ball-in-cup-catch ($961$ versus
$973$). It trails DreamerV3~\citep{hafner2025nature} on acrobot-swingup ($16$ versus
$69$) and finger-turn-hard ($579$ versus $616$). Across all three domains, these
results correspond to matching or exceeding the strongest reported baseline mean on
$15$ of $18$ tasks.

\subsection{Training dynamics across domains}
Figure~\ref{fig:curvegrid} shows how the endpoint comparisons in Table~\ref{tab:main}
emerge over training; black triangles mark the half- and full-budget scoring depths.
On several Meta-World~\citep{yu2020metaworld} tasks, endpoint convergence masks different routes to high
success. On box-close and sweep-into, \textsc{SDI} enters the near-saturated regime
after an initial learning phase and then remains stable. On hammer,
TD-MPC2~\citep{hansen2024tdmpc2} and R2-Dreamer~\citep{morihira2026r2} solve the task earlier, whereas \textsc{SDI} first plateaus near intermediate
success and makes a second, late transition to saturation. The ordering is reversed on
lever-pull: \textsc{SDI} rises rapidly to near-perfect success and maintains it while
the alternatives improve more slowly. On bin-picking and peg-insert-side,
R2-Dreamer~\citep{morihira2026r2}
makes earlier progress, but \textsc{SDI} continues improving and reaches the strongest
full-budget performance.

The precision-manipulation tasks show the clearest separation. \textsc{SDI} remains
near zero early on assembly and pick-out-of-hole, then undergoes a sustained late rise;
hand-insert improves more gradually across the second half of training. Disassemble is
different again: DreamerV3~\citep{hafner2025nature} improves first, but \textsc{SDI}'s final increase produces
the best endpoint. The Adroit~\citep{rajeswaran2018dapg} curves also distinguish early learning from eventual
performance. R2-Dreamer~\citep{morihira2026r2} reaches high hammer success fastest and \textsc{SDI} catches up
later, while \textsc{SDI} continues to gain on door and pen despite visible fluctuations
in the rolling success curves.

The DM-Control~\citep{tassa2018dmc} trajectories reflect the mixed endpoint pattern. Ball-in-cup-catch is
largely an early-saturation task for the strongest methods, whereas cartpole-swingup
and reacher-hard continue to separate methods later in training; on reacher-hard,
\textsc{SDI}'s advantage appears early and is largely maintained. By contrast,
DreamerV3~\citep{hafner2025nature} exhibits the strongest late-stage rise on acrobot-swingup and narrowly exceeds
\textsc{SDI} at the finger-turn-hard scoring depth. Thus, the suite-level averages
summarize task-dependent learning dynamics, including early breakthroughs,delayed transitions, and differences in plateau stability, rather than
uniform dominance throughout training.

\subsection{Ablation study}
\label{sec:mechres}
Table~\ref{tab:balance} isolates the contributions of the base agent, planning exposure,
action acceptance, exploration schedule, and replay prioritization on the three
precision-manipulation tasks. The complete method obtains a final equal-task mean of
$.53$ at $.97$ planner occupancy, compared with $.08$ for
R2-Dreamer~\citep{morihira2026r2} and $.05$ for the
DreamerV3~\citep{hafner2025nature} transfer variant.

At matched $.97$ occupancy, removing the acceptance gate reduces the three-task mean
from $.53$ to $.32$. The gate improves assembly from $.29$ to $.75$ and disassemble
from $.14$ to $.35$, while hand-insert changes from $.53$ to $.49$. Reducing planner
occupancy to $.47$ lowers the mean to $.16$ and leaves assembly and disassemble near
zero, indicating that sustained planning exposure is important in this setting.
The always-on bonus and reset controls reach means of $.06$ and $.08$, respectively.
Spread-prioritized replay improves hand-insert to $.40$ but does not transfer to
assembly or disassemble, yielding a mean of $.13$. Overall, no individual control
reproduces the full method's cross-task result.

\begin{figure}[t]\centering
\includegraphics[width=\linewidth]{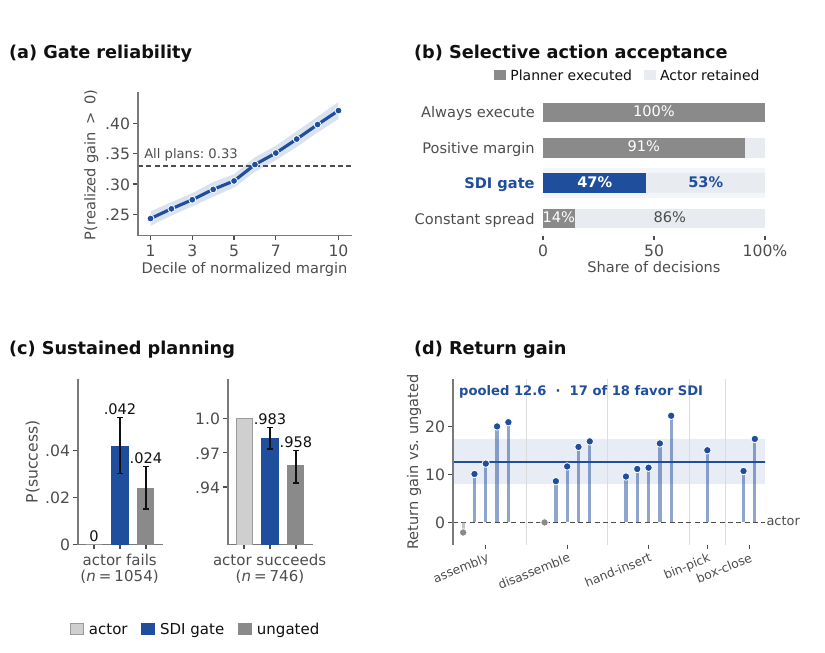}
\caption{\textbf{Paired-state planning experiments.}
(a) Realized one-decision improvement over the actor by normalized-margin decile;
the dashed line is the all-plan rate. (b) Proposal acceptance under four rules.
(c) Recovery and success preservation during $50$-step continuations from restored
states. (d) Checkpoint mean return difference between gated and ungated
continuations. Shading in (a) shows the standard error; intervals in (c,d) are $95\%$.}
\label{fig:calib}
\end{figure}

\subsection{Paired-state analysis}
\label{sec:findings}

\paragraph{Gate reliability and selectivity.}
\label{sec:cert}
For Figure~\ref{fig:calib}a, we sample $500$ decisions from each of $18$ checkpoints
across five Meta-World~\citep{yu2020metaworld} tasks and execute the actor and planned branches from the same
state. A decision is counted as beneficial only when the planned branch produces a strictly
positive realized return difference. This probability rises monotonically from about
$.24$ in the lowest normalized-margin decile to $.42$ in the highest, crossing the
all-plan rate of $.33$ in the upper half of the ranking. The imagined margin therefore
contains useful ordinal information about realized improvement. Figure~\ref{fig:calib}b shows the corresponding operating points. Always executing uses
every proposal, a positive-margin rule uses $91\%$, the \textsc{SDI} gate uses $47\%$,
and the constant-spread rule uses $14\%$. \textsc{SDI} thus occupies an intermediate
regime, rejecting a slight majority of proposals without becoming as restrictive as
the constant-spread rule. These execution shares measure selectivity rather than quality;
comparisons among rules must also account for which decisions each rule retains.

\paragraph{Recovery and success preservation.}
The sustained-planning study in Figure~\ref{fig:calib}c uses $1{,}800$ restored states
with planning available throughout each $50$-step continuation. Among the $1{,}054$
states where the actor fails, gated planning recovers success in $4.2\%$ of cases,
compared with $2.4\%$ for ungated planning: $44$ versus $25$ recoveries. Among the $746$
states where the actor succeeds, the gate preserves $98.3\%$ of successes, compared
with $95.8\%$ under ungated planning, corresponding to $13$ versus $31$ lost successes.
Selective execution therefore improves both sides of the intervention tradeoff: it
recovers more actor failures while damaging fewer actor successes. The pooled net
success change is $+.0172$ for gated planning, compared with $-.0033$ for ungated
planning. Figure~\ref{fig:calib}d directly compares the two planning rules at each checkpoint.

\section{Conclusion}
We introduced Sibling Disagreement in Imagination (\textsc{SDI}), which uses repeated
actor rollouts as a shared reference for task-directed exploration and selective
planning. The same actor-referenced comparison measures variation across imagined
alternatives and determines when a plan should replace the actor, without requiring
an additional disagreement ensemble. Planning guides data collection during training,
while evaluation retains the learned actor alone. Thus, the benchmark endpoints assess
policy improvement, while the paired-state experiments separately test the quality of
the planner's training-time interventions. Across $18$ tasks in three domains, \textsc{SDI} matches or exceeds the strongest
reported baseline on $15$ tasks. 
Overall, the findings support actor-referenced comparison as a practical way to
connect exploration and planning in world-model learning.

\label{endofbody}

\clearpage
\section*{AI Use Statement}
Generative AI tools assisted with methodological and experimental feedback; code
scaffolding and experiment orchestration; analysis scripts and interpretation checks;
literature search and bibliography verification; scientific-figure drafting; and
manuscript organization, title development, and language editing. The tools did not
generate synthetic datasets or autonomously collect experimental data, and this work
makes no theoretical-proof claims. The authors ran and inspected all AI-assisted code,
checked analyses and numerical claims against the retained experiment records, verified
citations against primary bibliographic sources, and manually reviewed all generated
text and figures. The authors take responsibility for the final content of this work,
including text, claims, code, and artifacts produced with AI assistance.

\section*{Ethics Statement}
This work studies model-based reinforcement learning in simulated control benchmarks
and does not involve human participants, animal subjects, personal or sensitive data,
or deployment on physical systems. Although the method is intended for general-purpose
control research, improved planning and autonomous decision making could be misapplied
in safety-critical or harmful settings. The reported benchmark results do not establish
safety, robustness, or suitability for real-world deployment; in particular, world-model
bias and planning error remain important limitations. Any transfer to physical or
high-stakes domains should therefore include domain-specific risk assessment, human
oversight, fail-safe constraints, and testing under distribution shift. We provide
code, configurations, and methodological details to support scrutiny and responsible
reuse.

\section*{Reproducibility Statement}
Section~\ref{sec:method} specifies the return construction, exploration bonus,
acceptance rule, planning schedule, and search settings. Section~\ref{sec:setup} defines
the task budgets, endpoints, and replication used for the main claims. Evaluation always uses the learned actor without planning.
The supplementary material contains only results with complete reported cohorts. Code,
configurations, and instructions are provided in the anonymous repository \url{https://anonymous.4open.science/r/sdi-EEDA}.

\clearpage
\bibliography{refs_submission}
\bibliographystyle{iclr2027_conference}

\clearpage
\appendix
\renewcommand{\thetable}{A\arabic{table}}
\renewcommand{\thefigure}{A\arabic{figure}}
\renewcommand{\theHtable}{appendix.\arabic{table}}
\renewcommand{\theHfigure}{appendix.\arabic{figure}}
\setcounter{table}{0}\setcounter{figure}{0}
\raggedbottom

\phantomsection
\section*{Limitations}
\label{sec:limits}

\paragraph{What the actor reference measures.}
Repeated actor rollouts quantify stochastic return variation for one fixed imagined
sequence; they do not capture model bias shared by all rollouts. The acceptance
threshold $\kappa$ is an empirical operating point rather than a statistical confidence
level. Restored-state experiments show that the normalized margin ranks the frequency
of realized improvement, but even its highest decile improves on the actor in fewer
than half of paired decisions. The score should therefore not be interpreted as a
calibrated probability, despite the better aggregate outcomes of gated than ungated
continuations.

\paragraph{Planning cost and simulator access.}
SDI reuses the base world model, actor, and critic, but search and fresh re-scoring add
acting-time computation even when a proposal is rejected. The paired-state analysis
also requires simulator-state restoration and common random numbers; these are analysis
tools and are not used during training. The experiments measure finite continuations
from frozen checkpoints and do not isolate the long-term learning contribution of each
accepted or rejected intervention.

\section{Method details}
\label{app:planner}\label{app:config}

\subsection{Search and rescoring}
The planner samples $N=256$ candidate sequences over horizon $H=8$, with $I=4$
refinement iterations, $E=32$ elites, and $R=4$ independent stochastic rollouts per
candidate. Search temperature is $1.0$; initial and minimum action standard deviations
are $.3$ and $.05$. Actions are clipped to the environment's bounds.
At each iteration,
\begin{align}
 a^{(n)} &= \mathrm{clip}(\mu+\sigma\odot\epsilon^{(n)},a_{\min},a_{\max}),
 \qquad \epsilon^{(n)}\sim\mathcal N(0,I),\\
 \hat G^{(n)} &= \frac1R\sum_{j=1}^R G_j(a^{(n)}),\qquad
 \zeta^{(n)}=\frac{\hat G^{(n)}-\mathrm{mean}_n\hat G^{(n)}}
 {\mathrm{sd}_n\hat G^{(n)}},\label{eq:mppi-score}\\
 w^{(e)} &= \mathrm{softmax}_{e\in\mathrm{elites}}(\zeta^{(e)}/\lambda),\qquad
 \mu\leftarrow\sum_e w^{(e)}a^{(e)},\label{eq:mppi-w}\\
 \sigma^2 &\leftarrow\mathrm{clip}\!\left(\sum_e w^{(e)}(a^{(e)}-\mu)^2,
 \sigma_{\min}^2,\sigma_{\max}^2\right).\label{eq:mppi-update}
\end{align}
Scores are standardized separately for each environment. The initial mean follows the
actor's imagined mode sequence. On subsequent acting steps the previous sequence is
shifted forward; episode boundaries reset this initialization.

After search, the plan and the actor's mode sequence receive $R_g=\max(R,4)=4$ fresh
rollouts each. The gate compares their mean returns using the standard deviation of
those actor rollouts. It executes only the first planned action; rejection falls back
to an action sampled from the actor. The reference used for scoring is therefore the
actor's fixed mode sequence, whereas the training-time fallback is stochastic.
Evaluation uses the actor's mode alone.

\paragraph{Independent model noise.}\label{app:coupling}
Candidate rollouts, repeated actor rollouts, and fresh re-scoring all use independent
model noise. Exploration uses a separate batch with its own rollout settings. Sharing
the starting latent state does not imply sharing the model's random draws. In contrast,
the executed-alternative studies restore simulator state and use common random numbers
where supported. These are evaluation procedures, not additional simulator access
required by SDI during learning.

\end{document}